\documentclass[letterpaper, 10 pt, conference]{ieeeconf}  % Comment this line out if you need a4paper

\IEEEoverridecommandlockouts                              % This command is only needed if 
\usepackage{amsmath} % assumes amsmath package installed
\usepackage{amssymb}  
\usepackage{booktabs}
\usepackage{multirow}
\usepackage{float}
\usepackage{colortbl}
\usepackage{threeparttable}
\usepackage{graphicx}    

\usepackage{xcolor}
\definecolor{mybrown}{rgb}{0.65, 0.16, 0.16} 

\usepackage{adjustbox}
 
\usepackage{xcolor}

\usepackage{pgfplots}
\usepackage{subcaption} 
\pgfplotsset{compat=newest}

\usepackage{hyperref}
\usepackage{cleveref}

\title{\LARGE \bf BridgeTA: Bridging the Representation Gap in Knowledge Distillation via Teacher Assistant for Bird’s Eye View Map Segmentation}

\author{Beomjun Kim$^{1, 2}$, Suhan Woo$^{1}$, Sejong Heo$^{2}$, and Euntai Kim$^{1, 3, \dagger}$%
\thanks{$\dagger$ Corresponding author.}%
\thanks{$^{1}$B.\,Kim, S.\,Woo, and E.\,Kim are with Yonsei University, South Korea, {\tt\small \{kbj,wsh112,etkim\}@yonsei.ac.kr}.}%
\thanks{$^{2}$B.\,Kim and S.\,Heo are with Hyundai Motor Company, South Korea, {\tt\small \{kim.beomjun,sejong.heo\}@hyundai.com}.}%
\thanks{$^{3}$E.\,Kim is with Korea Institute of Science and Technology, South Korea.}%
}

\begin{document}

\maketitle
\thispagestyle{empty}
\pagestyle{empty}

% Abstract
\begin{abstract}
Bird’s Eye View (BEV) map segmentation is one of the most important and challenging tasks in autonomous driving. Camera-only approaches have drawn attention as cost-effective alternatives to LiDAR, but they still fall behind LiDAR-Camera (LC) fusion-based methods. Knowledge Distillation (KD) has been explored to narrow this gap, but existing methods mainly enlarge the student model by mimicking the teacher’s architecture, leading to higher inference cost. To address this issue, we introduce BridgeTA, a cost‑effective distillation framework to bridge the representation gap between LC fusion and Camera‑only models through a Teacher Assistant (TA) network while keeping the student’s architecture and inference cost unchanged. A lightweight TA network combines the BEV representations of the teacher and student, creating a shared latent space that serves as an intermediate representation. To ground the framework theoretically, we derive a distillation loss using Young’s inequality, which decomposes the direct teacher-student distillation path into teacher-TA and TA-student dual paths, stabilizing optimization and strengthening knowledge transfer. Extensive experiments on the challenging nuScenes dataset demonstrate the effectiveness of our method, achieving an improvement of 4.2\% mIoU over the Camera-only baseline, up to 45\% higher than the improvement of other state-of-the-art KD methods.
The code will be available at \href{https://github.com/kxxbeomjun/BridgeTA}{https://github.com/kxxbeomjun/BridgeTA}.
\end{abstract}

%%%%%%%%%%%%%%%%%%%%%%%%%%%%%%%%%%%%%%%%%%%%%%%%%%%%%%%%%%%%%%%%%%%%%%%%%%%%%%%%

\section{Introduction}
Bird’s Eye View (BEV) map segmentation is a fundamental task in perception~\cite{li2024bevformer, liu2023bevfusion, peng2023bevsegformer, pan2023baeformer}, playing a crucial role in autonomous driving~\cite{ren2022collaborative, zhao2024bev}. By analyzing road components from a top-down perspective, BEV map segmentation provides essential spatial understanding, which is critical for ensuring the safety of autonomous vehicles. To balance cost and efficiency, recent research has increasingly focused on Camera-only methods~\cite{li2024bevformer, liu2023bevfusion, philion2020lift, zhou2022cross, zhu2023mapprior}, achieving notable performance improvements. However, despite these advancements, Camera-only methods still lag behind LiDAR-Camera fusion-based approaches in terms of performance~\cite{liu2023bevfusion, kim2024broadbev, le2024diffuser}.

\begin{figure}[t]
\centering
\begin{subfigure}[t]{0.44\textwidth}
    \centering
    \includegraphics[width=\textwidth]{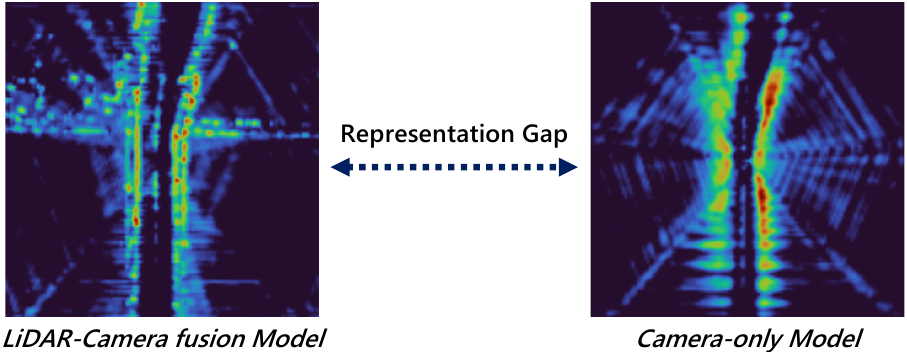}
    \caption{Visualization of BEV feature representation gap}
    \label{fig:figure1-1}
\end{subfigure}

\begin{subfigure}[t]{0.44\textwidth}
    \centering
    \includegraphics[width=\textwidth]{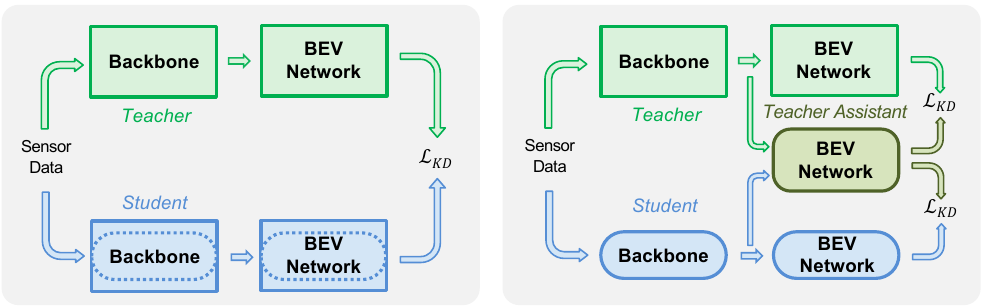}
    \caption{Existing KD methods vs. {\bf BridgeTA}}
    \label{fig:figure1-2}
\end{subfigure}
\caption{Due to fundamental differences in sensor modalities, the LiDAR-Camera fusion model and Camera-only model capture distinct information for semantic map segmentation, leading to the representation gap shown in (a). To cost-effectively address this gap, we propose BridgeTA, as illustrated in (b).}
\label{fig:figure1}
\end{figure}

To bridge this performance gap, BEV-based Knowledge Distillation (KD) methods, which distill knowledge from a fusion-based teacher to a camera-only student, have been attracting significant attention~\cite{wang2023distillbev, chen2022bevdistill, zhou2023unidistill}. In these methods, the representation gap due to the different input modalities between the teacher and student, as visualized in Figure~\ref{fig:figure1-1}, serves as a major obstacle. Existing approaches attempt to reduce this gap by designing the student model to follow the structure of the teacher model~\cite{zhao2024simdistill, hao2024mapdistill}, as depicted schematically in Figure~\ref{fig:figure1-2}. However, such methods increase the computational burden on the student, undermining the main advantage of KD, which is to achieve strong performance with a compact student. Additionally, many methods~\cite{zhou2023unidistill, li2022unifying} perform suboptimal distillation simply by forcing the student to replicate the representation of the teacher without adequately considering the inherent differences between them.

In this paper, we propose {\bf BridgeTA}, a novel distillation framework to {\bf Bridge} the representation gap through an innovative {\bf T}eacher {\bf A}ssistant network. Our TA structure can be applied without any modification to the student model and is used only during training. Consequently, it introduces no additional inference cost. Uniquely, our TA is constructed by combining the representations of the teacher and student, requiring no individual data input or dedicated backbone for the TA itself. Unlike previous TA-based KD methods~\cite{mirzadeh2020ta, liu2025monotakd}, this design makes our training process highly cost-efficient. By leveraging TA, we decompose the direct teacher-student distillation path into teacher-TA and TA-student dual paths, thereby alleviating the severe representation mismatch between the teacher and the student. To theoretically support our framework, we design our distillation loss based on \textit{Young’s inequality}~
\cite{hardy1952inequalities}. Furthermore, to maximize the effect of distillation, we also perform multi-level distillation through three distinct schemes. Rather than merely mimicking representations, our approach improves distillation efficiency by explicitly transferring the relative geometric relationships essential for encoding BEV representations. This enables the student to effectively capture crucial spatial dependencies.

Extensive experiments on the nuScenes dataset~\cite{caesar2020nuscenes} demonstrate that BridgeTA enhances BEV representation for BEV map segmentation while maintaining a cost-effective Camera-only setting during inference. Notably, BridgeTA decomposes the single distillation path into dual paths and achieves superior performance, significantly narrowing the gap between Camera-only and fusion-based models without incurring any additional cost or latency.

To summarize, our main contributions are as follows:
% \vspace{-1mm}
\begin{itemize}
  \setlength{\itemsep}{1pt}
  \setlength{\parskip}{0pt}
  \setlength{\parsep}{0pt}
    \item We introduce a novel TA-based distillation framework that bridges the representation gap between LC fusion teachers and Camera-only students without changing the student architecture or inference cost.
    \item We theoretically ground our dual-path distillation loss with \textit{Young’s inequality}, providing a tight upper bound for stable and effective knowledge transfer.
    \item We design a multi-level distillation scheme and conduct extensive experiments on the nuScenes dataset~\cite{caesar2020nuscenes}, where BridgeTA achieves a 4.2\% mIoU improvement over the Camera-only baseline, up to 45\% higher than the gains of other state-of-the-art KD methods.
\end{itemize}

%%%%%%%%%%%%%%%%%%%%%%%%%%%%%%%%%%%%%%%%%%%%%%%%%%%%%%%%%%%%%%%%%%%%%%%%%%%%%%%%
\section{Related Work}
\subsubsection{BEV map segmentation} Recently, BEV map segmentation has attracted significant attention in autonomous driving perception. Fusion-based methods combine LiDAR’s precise geometric and 3D spatial information with camera inputs to deliver superior performance and can be categorized according to the fusion stage, such as early fusion~\cite{sindagi2019early1}, deep fusion~\cite{liu2023bevfusion, kim2024broadbev}, and late fusion~\cite{bai2022transfusion, chen2023futr3d}. While these approaches enhance BEV representations, they introduce high computational overhead and LiDAR dependency costs. As a result, Camera-only methods~\cite{chen2024residual, woo2024location} have recently attracted increasing attention. To improve BEV representation and prediction performance, Camera-only approaches have adopted advanced techniques such as attention mechanisms~\cite{zhou2022cross, chen2022gkt, li2024bevformer} and diffusion models~\cite{ho2020ddpm}. However, the increasing complexity of these methods often leads to higher inference latency, diminishing their efficiency advantages over fusion-based methods. To overcome this limitation, we propose a novel distillation framework that improves the performance of Camera-only methods without additional inference cost or complexity.

\begin{figure*}
  \centering
    \includegraphics[width=0.97\linewidth]{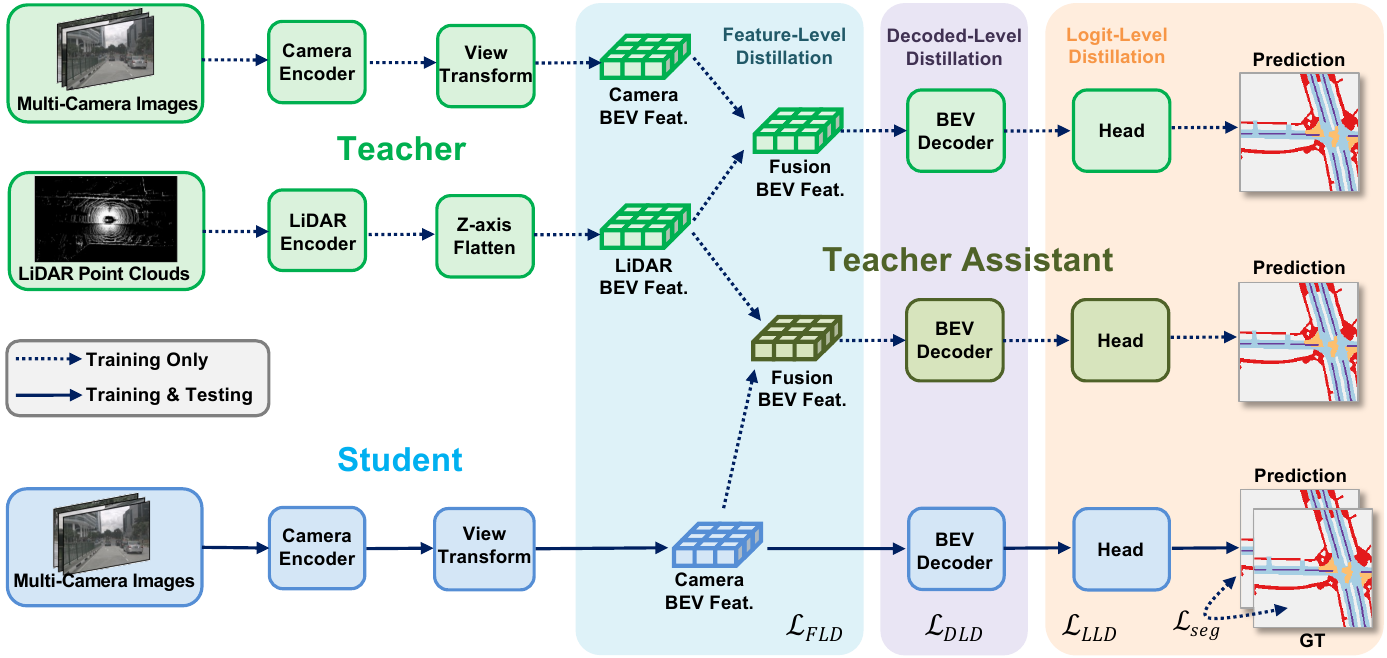}
  \caption{ {\bf Overview of BridgeTA framework.} We propose a novel distillation framework to bridge the representation gap between LiDAR-Camera fusion teacher (green) and Camera-only student (blue) through a Teacher Assistant network (dark green). We design a multi-level distillation framework with three distinct distillation modules for the student model to learn rich BEV representation from the teacher model through an efficient KD process. The teacher model and TA network are used only during training, ensuring that inference relies solely on the Camera, without LiDAR.}
  \label{fig:overallframework}
\end{figure*}

\subsubsection{Knowledge Distillation (KD)} Initially, KD was introduced as an effective approach for transferring knowledge from a high-capacity teacher to a lightweight student for model compression~\cite{hinton2015distilling}. Then, it has been extended to vision tasks such as semantic segmentation~\cite{yang2022semantic} and object detection~\cite{wang2023distillbev}. While early KD methods focused primarily on logit distillation~\cite{zhao2022logit2}, recent studies have explored feature distillation~\cite{heo2019feature1, liu2020residual} to provide richer guidance. However, KD methods often struggle to close representation and performance gaps between teacher and student when these gaps are large. Some approaches have addressed this by introducing a teacher assistant (TA) to mediate distillation, but existing TA-based methods~\cite{mirzadeh2020ta, liu2025monotakd} typically require extra data inputs and a dedicated backbone for the TA, resulting in inefficient, resource-intensive training. Other works have tried to reduce the gap by aligning the student’s structure with the teacher’s~\cite{zhao2024simdistill, hao2024mapdistill}, but this usually incurs significant inference cost. In this work, we propose a cost-effective TA-based distillation framework that overcomes these limitations by efficiently bridging the teacher-student gap without increasing training or inference complexity.

%%%%%%%%%%%%%%%%%%%%%%%%%%%%%%%%%%%%%%%%%%%%%%%%%%%%%%%%%%%%%%%%%%%%%%%%%%%%%%%%
\section{Method}
In this section, we detail the novelty and cost-effectiveness of the BridgeTA framework. We first present the architectures of the teacher, TA, and student models. Next, we introduce the distillation path decomposition method leveraging Young’s inequality, and subsequently provide detailed descriptions of each level in the proposed multi-level distillation scheme. An overview of BridgeTA is shown in Figure~\ref{fig:overallframework}.

\subsection{Model architecture}

\subsubsection{Teacher and Student} BridgeTA is developed based on the BEVFusion~\cite{liu2023bevfusion} codebase, which is widely used for BEV map segmentation tasks. The teacher model utilizes both the LiDAR and Camera branches, while the student uses only the Camera branch. To enable effective knowledge transfer, we align the teacher’s channel dimensions and decoder with the student’s structure. Unlike previous methods~\cite{zhao2024simdistill,hao2024mapdistill} that modify the student architecture, our approach fully preserves the original student model.

\subsubsection{Teacher Assistant (TA)} We propose a novel TA-based distillation framework to address both the representation and performance gaps in BEV map segmentation. BridgeTA introduces a TA network that requires no individual data input or dedicated backbone, fusing the teacher’s LiDAR BEV feature with the student’s Camera BEV feature, as in Figure~\ref{fig:overallframework}. The TA shares BEV decoder and head architecture with both teacher and student for consistency. This structure decomposes the direct teacher-to-student path into teacher-to-TA and TA-to-student. As a result, the student receives finer-grained and less divergent guidance, rather than a single large distillation signal from the teacher. Therefore, the training process becomes more stable and effective, allowing better alignment.

\subsection{Distillation Loss Formulation}

Our framework, which decomposes the direct distillation path into dual-paths, is theoretically grounded by recalling the classical \textit{Young’s inequality}~\cite{hardy1952inequalities}, which states that for any $a, b \in \mathbb{R}$ and $\lambda > 0$,
\begin{equation}
|ab| \leq \frac{\lambda}{2} a^2 + \frac{1}{2\lambda} b^2,
\label{eq:young-scalar}
\end{equation}

\noindent this inequality extends to vectors or matrices using the L2 norm or Frobenius norm, respectively. Specifically, for any $X$, $Y$ and $\lambda > 0$,
\begin{equation}
    2\langle X, Y \rangle \leq \lambda \|X\|^2 + \frac{1}{\lambda} \|Y\|^2.
    \label{eq:young-inner}
\end{equation}

\noindent Expanding the squared norm of a sum gives:
\begin{equation}
\|X + Y\|^2 = \|X\|^2 + 2\langle X, Y \rangle + \|Y\|^2,
\label{eq:norm-expansion}
\end{equation}

\noindent and by applying Eq.~\eqref{eq:young-inner} to the cross-term in Eq.~\eqref{eq:norm-expansion} and then setting $\lambda = \varepsilon$ leads to:
\begin{equation}
\|X + Y\|^2 \leq (1 + \varepsilon) \cdot \|X\|^2 + \left(1 + \frac{1}{\varepsilon}\right) \cdot \|Y\|^2.
\label{eq:young-gen}
\end{equation}

\noindent Finally, substituting $X = R^S - R^{TA}$, $Y = R^{TA} -R^T$ into Eq.~\eqref{eq:young-gen}, we obtain the following inequality:
\begin{equation}
\label{eq:young}
\begin{split}
\| R^S - R^T \|^2 \leq\; & (1+\varepsilon) \cdot \| R^S - R^{TA} \|^2 \\
& + \left(1 + \frac{1}{\varepsilon}\right) \cdot \| R^{TA} - R^T \|^2,
\end{split}
\end{equation}

\noindent where $R^{S}$, $R^{T}$, and $R^{TA}$ denote the student, teacher, and TA representations at each distillation level, respectively.

Here, the left-hand side represents the direct teacher-to-student distillation, while the right-hand side is our proposed dual-path distillation loss. If the sum of the teacher-to-TA and TA-to-student losses converges, the upper bound ensures that the student and teacher representations also become close. In other words, effective optimization of our dual-path objective leads to successful alignment of the student with the teacher.

Furthermore, to achieve the tightest upper bound of the inequality, we define the following objective function, Eq.~\eqref{eq:young_ab} shown below: 

\begin{equation}
\label{eq:young_ab}
    f(\varepsilon) = (1 + \varepsilon) \cdot a^2 + \left(1 + \frac{1} {\varepsilon}\right) \cdot b^2,
\end{equation}

\noindent where $a = \|R^S - R^{TA}\|$, $b = \|R^{TA} -R^T\|$, and $\varepsilon > 0$. To find the optimal value, we take the derivative of $f(\varepsilon)$ with respect to $\varepsilon$ and solve for zero, yielding
\begin{equation}
    \label{eq:young_extremum}
    \frac{df}{d\varepsilon}
    = a^2 \;-\; \frac{b^2}{\varepsilon^2}
    = 0
    \quad\Longrightarrow\quad
    \varepsilon^* = \frac{b}{a}\,.
\end{equation}
\noindent Here, $\varepsilon^*$ is an optimal value which achieves the tightest possible upper bound for our decomposition. Setting $\varepsilon = \varepsilon^*$ ensures a strong theoretical guarantee that the teacher-student representation gap will be effectively bridged through our TA-based distillation framework.

\subsection{Multi-Level Knowledge Distillation}

\begin{figure}[t]
  \centering
   \includegraphics[width=0.95\linewidth]{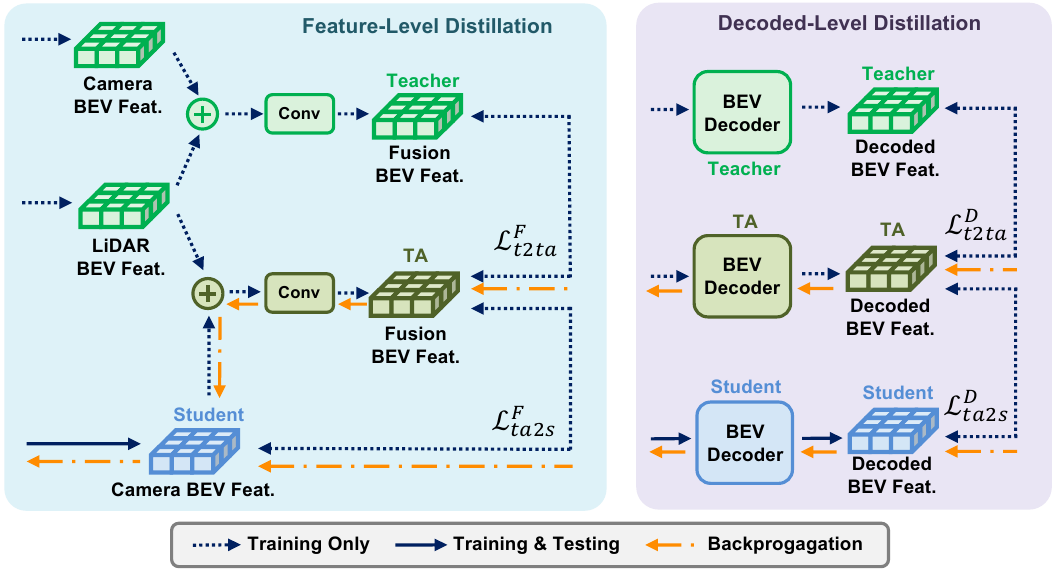}
   \caption{Illustration of {\bf Feature-Level Distillation (FLD)} and {\bf Decoded-Level Distillation (DLD)}. Both FLD and DLD utilize dual-path distillation with the TA network to bridge the BEV representation gap between the teacher and student models.}
   \label{fig:fld}
\end{figure}

To maximize distillation effectiveness in bridging both the representation and performance gaps, we propose a multi-level distillation framework comprising three schemes: Feature-Level Distillation (FLD), Decoded-Level Distillation (DLD), and Logit-Level Distillation (LLD). The primary advantage of multi-level distillation is that it captures complementary information across network stages, enabling comprehensive knowledge transfer. These three schemes share a common concept of a dual distillation path between the teacher-TA and TA-student, as illustrated in Figure~\ref{fig:fld} and Figure~\ref{fig:logit}. We employ mean square error (MSE) between corresponding representations to unify the loss formulation across all levels. Specifically, the teacher-to-TA and TA-to-student objectives are defined as follows:
\begin{equation}
\begin{split}
    \mathcal{L}_{t2ta}^{X} = \frac{1}{H_X \times W_X} \sum_{i}^{H_X} \sum_{j}^{W_X} \left\| (R_{X}^{TA})_{i,j} - (R_{X}^{T})_{i,j} \right\|^2, \\
    \mathcal{L}_{ta2s}^{X} = \frac{1}{H_X \times W_X} \sum_{i}^{H_X} \sum_{j}^{W_X} \left\| (R_{X}^{S})_{i,j} - (R_{X}^{TA})_{i,j} \right\|^2,
\end{split}
\label{eq:general_dual_loss}
\end{equation}

\noindent where $R_{X}^{T}$, $R_{X}^{TA}$, and $R_{X}^{S}$ denote the teacher, TA, and student representations at each distillation level $X$. In addition, $H_X$, $W_X$ represent the corresponding spatial dimensions. The precise forms of these variables depend on the chosen distillation level (e.g., feature, decoded, or logit), and are specified in the respective subsections below.

\subsubsection{Feature-Level Distillation (FLD)} At the feature level, the representation gap between the teacher and student arises from their different input modalities: the teacher captures both geometric information from LiDAR and semantic texture from Cameras, while the student relies only on Camera inputs. To bridge this gap, we propose distillation at the BEV feature level using a TA network that fuses the teacher’s LiDAR BEV features with the student’s Camera BEV features, thereby forming an intermediate representation that integrates both geometric richness and semantic information. Through this process, the student not only learns from the teacher’s LiDAR-enhanced representation, but also implicitly adapts its own Camera features to better align with the LiDAR modality for more effective fusion. 

For the FLD stage, we instantiate the general loss formulation in Eq.~\eqref{eq:general_dual_loss} by specifying the variables as follows:
\[
(R_{X}^{T},\, R_{X}^{TA},\, R_{X}^{S}) = (F_{fus}^{T},\, F_{fus}^{TA},\, F_{cam}^{S}),
\]
where $F_{fus}^{T}$ and $F_{fus}^{TA}$ are the fused BEV features of the teacher and TA, and $F_{cam}^{S}$ is the Camera BEV feature of the student. The spatial size is $H_f \times W_f$. We define these features as ${F}_{fus}^{T} \in \mathbb{B}^{{C}_{f}^{T} \times H_{f} \times W_{f}}$, ${F}_{fus}^{TA} \in \mathbb{B}^{{C}_{f}^{TA} \times H_{f} \times W_{f}}$, and ${F}_{cam}^{S} \in \mathbb{B}^{{C}_{c}^{S} \times H_{f} \times W_{f}}$. To facilitate distillation, we set the channel and spatial dimensions to be identical, i.e., ${C}_{f}^{T} = {C}_{f}^{TA} = {C}_{c}^{S}$ and $H_{f} \times W_{f}$. The resulting feature-level distillation loss is then formulated as
\begin{equation}
    \mathcal{L}_{\mathrm{FLD}} = (1 + \varepsilon^*) \cdot \mathcal{L}_{ta2s}^{F} + \left(1 + \frac{1}{\varepsilon^*}\right) \cdot \mathcal{L}_{t2ta}^{F},
\end{equation}
where $\varepsilon^*$ is optimally chosen as discussed previously.

\subsubsection{Decoded-Level Distillation (DLD)}
The decoded level serves as an intermediate stage between the feature level and the prediction head, playing a vital role in refining high-level scene understanding and spatial relationships. Distilling at this stage enables the transfer of more structured and semantically rich information, thereby helping the student model to interpret complex scenes more accurately.

For the DLD stage, we instantiate the general dual-path loss formulation in Eq.~\eqref{eq:general_dual_loss} by specifying as:
\[
(R_X^{T},\, R_X^{TA},\, R_X^{S}) = (F_{dec}^{T},\, F_{dec}^{TA},\, F_{dec}^{S}),
\]
where $F_{dec}^{T}$ and $F_{dec}^{TA}$ denote the decoded features of the teacher and TA networks, and $F_{dec}^{S}$ is the decoded feature of the student. The spatial size at this stage is $H_d \times W_d$. All decoded features are defined as ${F}_{dec}^{T}, {F}_{dec}^{TA}, {F}_{dec}^{S} \in \mathbb{B}^{{C}_{d} \times H_{d} \times W_{d}}$, where ${C}_{d}$, $H_{d}$, and $ W_{d}$ denote the shared channel and spatial dimensions. The resulting decoded-level distillation loss is then formulated as
\begin{equation}
    \mathcal{L}_{\mathrm{DLD}} = (1 + \varepsilon^*) \cdot \mathcal{L}_{ta2s}^{D} + \left(1 + \frac{1}{\varepsilon^*}\right) \cdot \mathcal{L}_{t2ta}^{D},
\end{equation}
where $\varepsilon^*$ is optimally chosen as described previously.

\subsubsection{Logit-Level Distillation (LLD)} At the logit level, distillation directly influences model performance by aligning the predictions of teacher and student. To minimize representation and performance gaps, we extend the dual-path KD approach to the logit level using the TA network, enabling the student to better capture the teacher’s logit distributions.

For the LLD stage, we instantiate Eq.~\eqref{eq:general_dual_loss} with
\[
(R_X^{T},\, R_X^{TA},\, R_X^{S}) = (L^{T_T},\, L^{TA_{TA}},\, L^{S_S}),
\]
where $L^{T_T}$ and $L^{TA_{TA}}$ are the logit outputs of the teacher and TA heads (given their respective features), and $L^{S_S}$ is the student’s logit output. The spatial size at this stage is $H \times W$. The base loss of LLD module is formulated as:
\begin{equation}
    \mathcal{L}_{Base}^L = (1 + \varepsilon^*) \cdot \mathcal{L}_{ta2s}^{B} + \left(1 + \frac{1} {\varepsilon^*}\right) \cdot \mathcal{L}_{t2ta}^{B},
\end{equation}
\noindent where $\varepsilon^*$ is chosen as in the FLD and DLD losses.

\begin{figure}[t]
  \centering
   \includegraphics[width=0.95\linewidth]{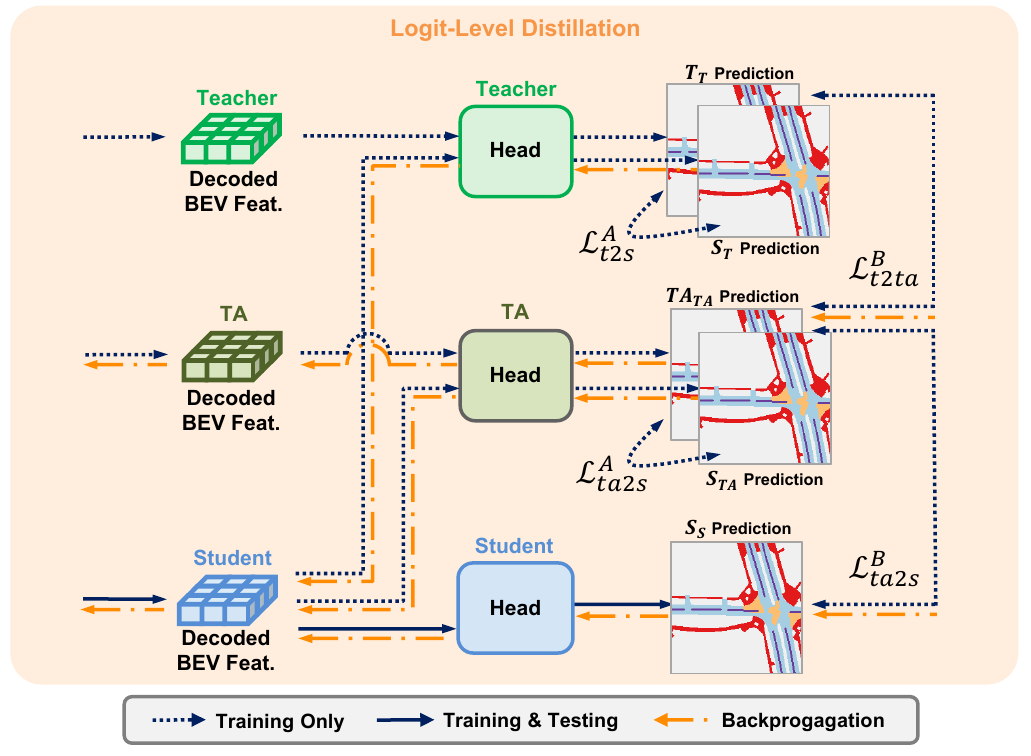}
   \caption{Illustration of {\bf Logit-Level Distillation (LLD)}. LLD utilizes multi-path distillation to effectively bridge both logit representation and performance gaps between the teacher and student models, with four distinct loss terms that provide richer guidance and accelerate training convergence.}
   \label{fig:logit}
\end{figure}

We further enhance logit-level distillation with a structure inspired by CrossKD~\cite{wang2024crosskd}. In this setup, the decoded BEV representation from the student is passed not only to the student’s Head but also to the teacher’s and TA’s Heads, generating $S_S$, $S_T$, and $S_{TA}$ predictions, as shown in Figure~\ref{fig:logit}. This design enriches the distillation paths and enables more comprehensive knowledge transfer, as the well-trained heads of the teacher and TA provide the student with diverse and informative guidance.

By applying each head to both its own feature and the student’s feature, we obtain four logit outputs: $L^{T_T}$, $L^{TA_{TA}}$, $L^{S_T}$, and $L^{S_{TA}}$. This configuration allows us to employ KL divergence as an auxiliary distillation loss between outputs of the same head under different inputs, as shown below:
\begin{align}
\begin{gathered}
    \mathcal{L}_{Aux}^{L} = \mathcal{L}_{t2s}^{A} + \mathcal{L}_{ta2s}^{A}, \\
    \mathcal{L}_{Aux}^{L} = KL(L^{T_T} \parallel L^{S_T}) + KL(L^{TA_{TA}} \parallel L^{S_{TA}}).
\end{gathered}
\end{align}

This additional supervision accelerates convergence and strengthens the student to learn robust and reliable logit representations, ultimately narrowing the performance gap. The overall logit-level distillation loss combines the dual-path regression and the auxiliary KL terms, leveraging both fine-grained regression and distribution-level alignment:
\begin{equation}
    \mathcal{L}_{LLD} = \mathcal{L}_{Base}^L + \mathcal{L}_{Aux}^{L}.
\end{equation}
Here, $L^{T_T}$, $L^{TA_{TA}}$, $L^{S_S}$, $L^{S_T}$, and $L^{S_{TA}}$ all denote logit outputs in $\mathbb{R}^{N_c \times H \times W}$, where $N_c$ is the number of classes and $H \times W$ is the BEV prediction map size. 

\subsection{Training and Inference}
\subsubsection{Training} We formulate BEV map segmentation as a pixel-wise classification problem and optimize the student model with a segmentation loss. The total loss used to train BridgeTA is:
\begin{equation}
    \mathcal{L}_{total} =  \mathcal{L}_{seg} + \lambda_1 \cdot \mathcal{L}_{FLD} + \lambda_2 \cdot \mathcal{L}_{DLD} + \lambda_3 \cdot \mathcal{L}_{LLD},
\end{equation}
\noindent where $\mathcal{L}_{\text{seg}}$ is the segmentation loss of the student model, which is applied to maximize the Intersection over Union (IoU) value. The hyperparameters $\lambda_1, \lambda_2$, and $\lambda_3$ are introduced to balance the contributions of the distillation losses,  $\mathcal{L}_{\text{FLD}}, \mathcal{L}_{\text{DLD}},$ and $\mathcal{L}_{\text{LLD}}$, respectively.

Moreover, the introduction of level-specific weights can be theoretically justified by Young’s inequality. For each distillation level $\ell$, the discrepancy between student and teacher representations can be bounded by the Teacher-TA and TA-Student terms. Extending this to all levels with weights $\lambda_\ell$, we obtain:
\begin{equation}
\begin{split}
\sum_\ell \lambda_\ell \| R^S_\ell - R^T_\ell \|^2 \leq\;
& \sum_\ell \lambda_\ell \Big[(1+\varepsilon_\ell) \| R^S_\ell - R^{TA}_\ell \|^2 \\
& \qquad + \left(1 + \frac{1}{\varepsilon_\ell}\right) \| R^{TA}_\ell - R^T_\ell \|^2\Big],
\end{split}
\label{eq:weighted-dual-path}
\end{equation}
\noindent where $\varepsilon_\ell > 0$ is a positive scalar. This formulation provides a principled explanation for introducing $\lambda_\ell$, aligning the total loss with a theoretically grounded upper bound.

\subsubsection{Inference} In our approach, we fully optimize the Camera-only student model through a novel distillation framework, enabling it to capture rich BEV representations comparable to those of the LC fusion-based teacher model. Furthermore, by leveraging a cost-effective distillation framework, we use only the Camera branch of the student model during the inference stage, excluding the teacher and TA networks at this stage, thereby avoiding any increase in computational cost or inference latency.

%%%%%%%%%%%%%%%%%%%%%%%%%%%%%%%%%%%%%%%%%%%%%%%%%%%%%%%%%%%%%%%%%%%%%%%%%%%%%%%%
\section{Experiments}
\begin{table*}[t]
    \centering
    \renewcommand{\arraystretch}{1.1}
    \setlength{\tabcolsep}{11pt} 
    \fontsize{9}{11}\selectfont   
    \begin{tabular}{lccccccc|c}
        \toprule
        \multirow{2}{*}{Method} & \multirow{2}{*}{Modality} & \multicolumn{7}{c}{IoU\(\uparrow\)(\%)} \\
        \cline{3-9}
         & & Drivable & Ped. Cross & Walkway & Stopline & Carpark & Divider & Mean \\
        \midrule
        SimDistill~\cite{zhao2024simdistill} & LC $\rightarrow$ C & 82.4 & 57.3 & 61.3 & 51.1 & 54.4 & 48.3 & 59.2 \\
        MapDistill~\cite{hao2024mapdistill} & LC $\rightarrow$ C & 82.6 & 57.4 & 61.7 & 51.6 & 54.5 & 48.8 & 59.5 \\
        \rowcolor{gray!14}
        \textbf{BridgeTA (Ours)} & LC $\rightarrow$ C & \textbf{83.3} & \textbf{58.6} & \textbf{62.9} & \textbf{53.6} & \textbf{56.6} & {\bf 50.1} & \textbf{60.8} \\
        \bottomrule
    \end{tabular}
    \caption{ Comparison against state-of-the-art KD methods. We compare BEV map segmentation performance on the nuScenes validation set. BridgeTA surpasses all other KD methods, achieving top performance across all classes. LC $\rightarrow$ C denotes distillation from a LiDAR-Camera fusion teacher model to a Camera-only student model.}
    \label{tab:comparekd}
\end{table*}

\begin{table*}[t]
    \centering
    \renewcommand{\arraystretch}{1.1}
    \setlength{\tabcolsep}{14pt}  
    \fontsize{9}{11}\selectfont 
    \begin{tabular}{lcccccc|c}
        \toprule
        \multirow{2}{*}{Method} & \multicolumn{7}{c}{IoU\(\uparrow\)(\%)} \\
        \cline{2-8}
         & Drivable & Ped. Cross & Walkway & Stopline & Carpark & Divider & Mean \\
        \midrule
        LSS~\cite{philion2020lift} & 75.4 & 38.8 & 46.3 & 30.3 & 39.1 & 36.5 & 44.4 \\
        CVT~\cite{zhou2022cross} & 74.3 & 36.8 & 39.9 & 25.8 & 35.0 & 29.4 & 40.2 \\
        M$^2$BEV~\cite{xie2022m2bev} & 77.2 & - & - & - & - & 40.5 & - \\
        BEVFusion-C~\cite{liu2023bevfusion} & 81.7 & 54.8 & 58.4 & 47.4 & 50.7 & 46.4 & 56.6 \\
        MapPrior~\cite{zhu2023mapprior} & 81.7 & 54.6 & 58.3 & 46.7 & 53.3 & 45.1 & 56.7 \\
        X-Align~\cite{borse2023x} & 82.4 & 55.6 & 59.3 & 49.6 & 53.8 & 47.4 & 58.0 \\
        MetaBEV~\cite{ge2023metabev} & 83.3 & 56.7 & 61.4 & 50.8 & 55.5 & 48.0 & 59.3 \\
        DDP~\cite{ji2023ddp} & {\bf 83.6} & 58.3 & 61.6 & 52.4 & 51.4 & 49.2 & 59.4 \\
        RGC~\cite{chen2024residual} & 81.7 & 57.1 & 60.5 & 51.7 & 53.8 & {\bf 53.5} & 59.7 \\
        \rowcolor{gray!14}
        \textbf{BridgeTA (Ours)} & 83.3 & \textbf{58.6} & \textbf{62.9} & \textbf{53.6} & \textbf{56.6} & 50.1 & \textbf{60.8} \\
        \bottomrule
    \end{tabular}
    \caption{Comparison against state-of-the-art Camera-only BEV map segmentation methods. We compare performance on the nuScenes validation set. BridgeTA achieves the highest mIoU overall and outperforms all other methods in 4 out of 6 classes.}
    \vspace{-5pt}
    \label{tab:comparebev}
\end{table*}

\subsection{Implementation Details}

\subsubsection{Dataset} To assess the effectiveness of our method for BEV map segmentation, we conduct experiments on the nuScenes~\cite{caesar2020nuscenes} dataset, a widely used benchmark. The dataset contains approximately 1.4M Camera images, 390K LiDAR sweeps, and HD maps covering 40K keyframes, collected in Boston and Singapore using a 32-channel LiDAR and six RGB Cameras. Annotations are provided for keyframes. nuScenes encompasses diverse challenging conditions, such as rain, night, and complex intersections, enabling a comprehensive evaluation of BridgeTA's robustness and effectiveness. We follow the experimental protocol of LSS~\cite{philion2020lift} and implement BridgeTA using MMDetection3D~\cite{mmdet3d2020}.

\subsubsection{Train setting} All experiments were conducted on 2 NVIDIA RTX A6000 GPUs with the teacher model frozen. The student model was trained for 20 epochs using a batch size of 6, a learning rate of 1e-4, and a Cosine Annealing schedule~\cite{loshchilov2016sgdr}.

\subsection{State-of-the-Art Comparison Results}
\subsubsection{KD methods} We compared BridgeTA with state-of-the-art (SOTA) KD methods in BEV networks, as shown in Table~\ref{tab:comparekd}. To ensure fairness, we re-implemented these methods on the BEVFusion~\cite{liu2023bevfusion} codebase to match our experimental settings. BridgeTA achieves higher accuracy across all classes, with up to 45\% over other KD methods (4.2 vs. 2.9 mIoU). Moreover, as shown in Table~\ref{tab:computationalcost}, since BridgeTA introduces no extra computational cost or latency over the baseline, it achieves up to 1.2× faster inference (16.2 vs. 13.3 FPS), 12\% lower memory (7.6 vs. 8.6 GB), 14\% fewer FLOPs (465.6 vs. 541.0 G), and 12\% fewer parameters (31.8 vs. 36.2 M) than other KD methods during inference. These results underscore the efficiency and effectiveness of our KD framework.

\subsubsection{Camera-only BEV map segmentation method} We also compare ours with state-of-the-art Camera-only BEV map segmentation methods. As shown in Table~\ref{tab:comparebev}, BridgeTA achieves the highest mIoU among them. Specifically, it improves performance by 4.2\% mIoU over the baseline, BEVFusion-C, demonstrating effective knowledge transfer.

\begin{table}[t]
    \centering
    \begingroup
    \setlength{\tabcolsep}{1.7mm}
    \renewcommand{\arraystretch}{1.1}
    \fontsize{9}{11}\selectfont  
    \begin{tabular}{lccc}
    \toprule
    Method & Latency (ms) & FLOPs (G) & Params (M) \\
    \midrule
    Baseline & 61.6 & 456.6 & 31.8  \\
    SimDistill~\cite{zhao2024simdistill} & 69.2(+7.6) & 493.5(+36.9) & 32.4(+0.6) \\
    MapDistill~\cite{hao2024mapdistill} & 75.2(+13.6) & 541.0(+84.4) & 36.2(+4.4) \\
    \rowcolor{gray!14} 
    {\bf BridgeTA (Ours)} & {\bf 61.6(+0.0)} & {\bf 456.6(+0.0)} & {\bf 31.8(+0.0)} \\
    \bottomrule
    \end{tabular}
    \endgroup
    \caption{Computational efficiency comparison against state-of-the-art KD methods. Unlike other KD methods, BridgeTA incurs no additional computation cost or latency.}
    \vspace{-10pt}
    \label{tab:computationalcost} 
\end{table}

\begin{figure*}[ht]
    \centering
    \begin{subfigure}[t]{0.3\textwidth}
        \centering
        \caption{Avg. $L2$ distance: $F_{fus}^{T}$ vs $F_{cam}^{S}$}
        \includegraphics[width=\textwidth]{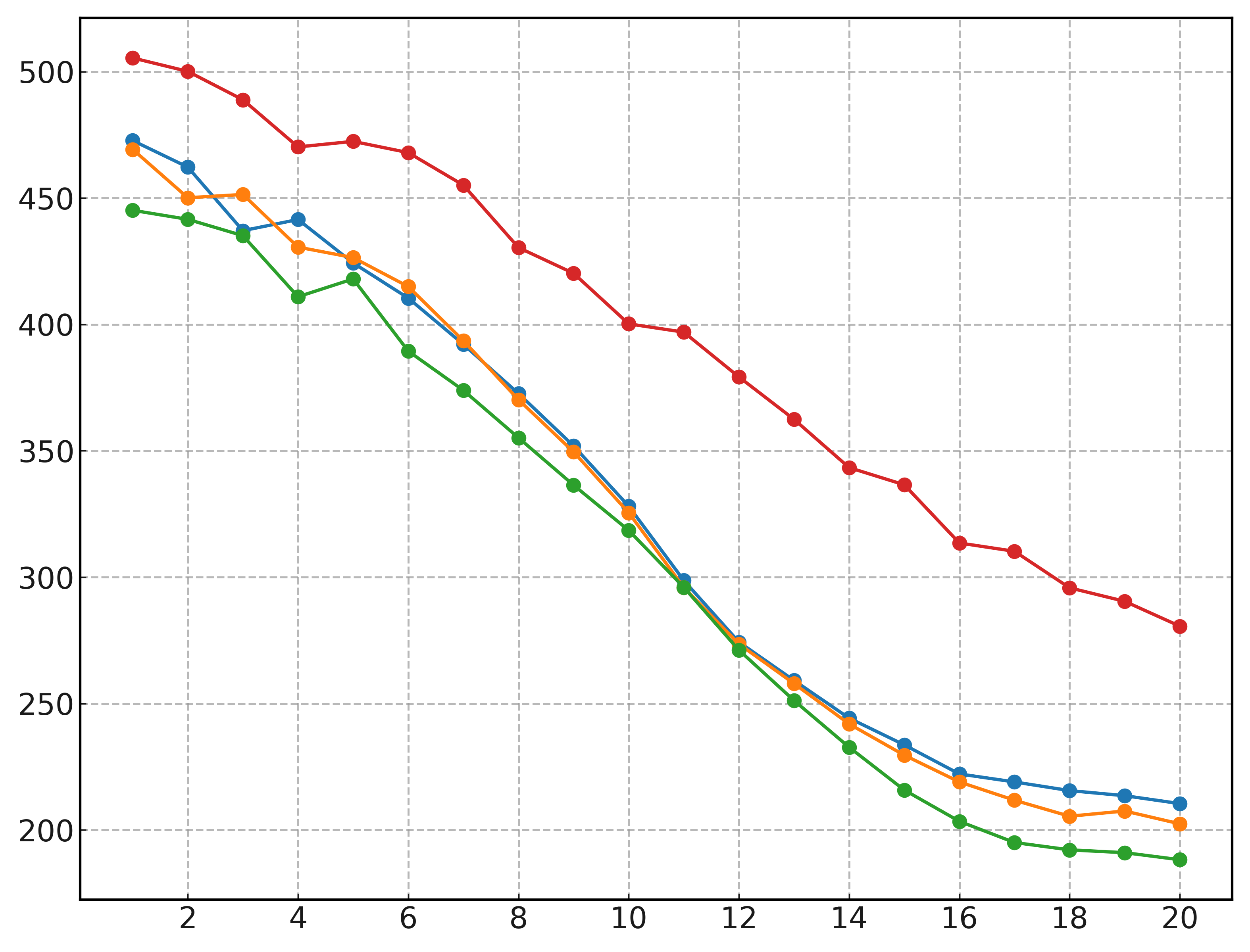}
    \end{subfigure}
    \hfill
    \begin{subfigure}[t]{0.3\textwidth}
        \centering
        \caption{Avg. $L2$ distance: $F_{dec}^{T}$ vs $F_{dec}^{S}$}
        \includegraphics[width=\textwidth]{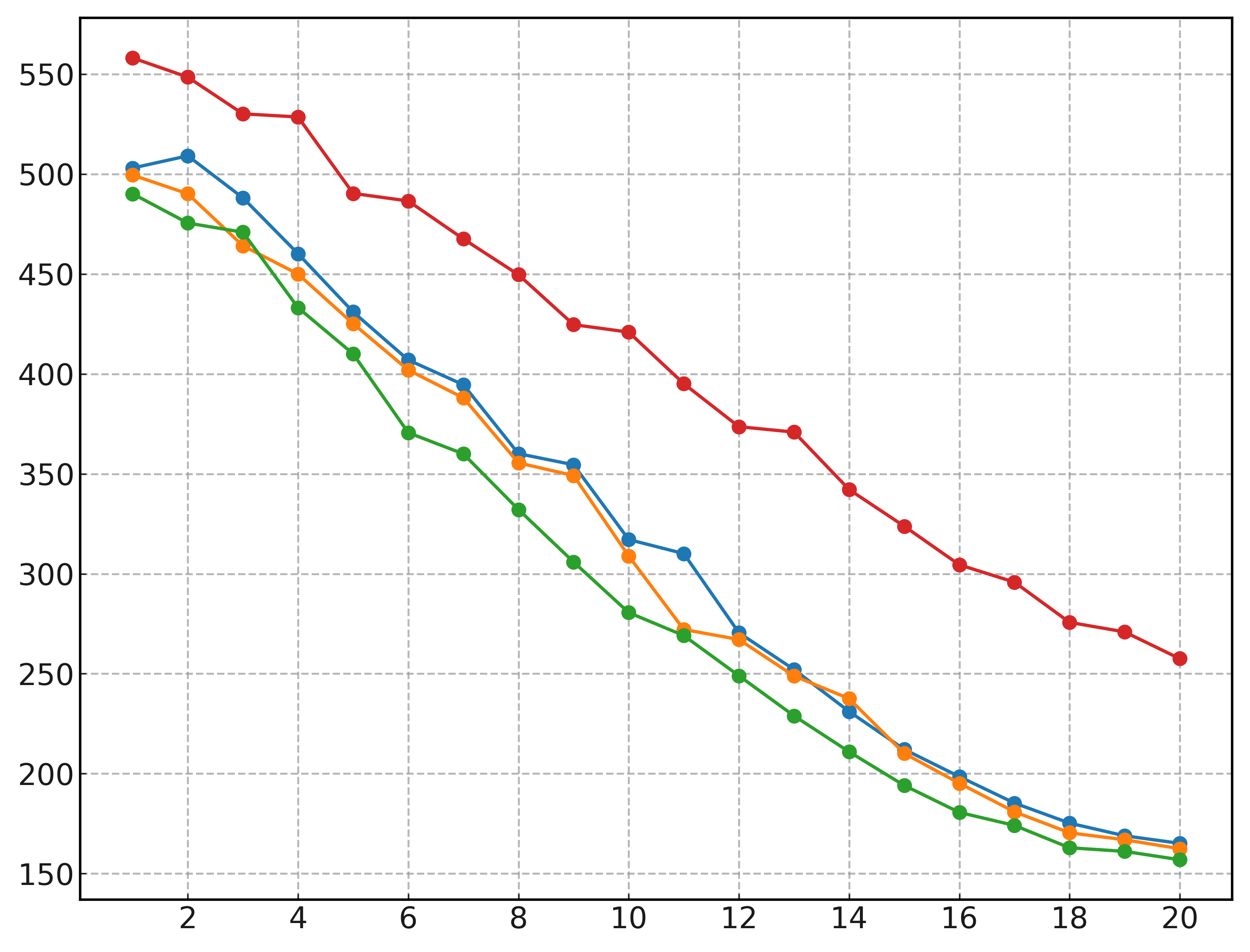}
    \end{subfigure}
    \hfill
    \begin{subfigure}[t]{0.3\textwidth}
        \centering
        \caption{Avg. $L2$ distance: $L^{T_T}$ vs $L^{S_S}$}
        \includegraphics[width=\textwidth]{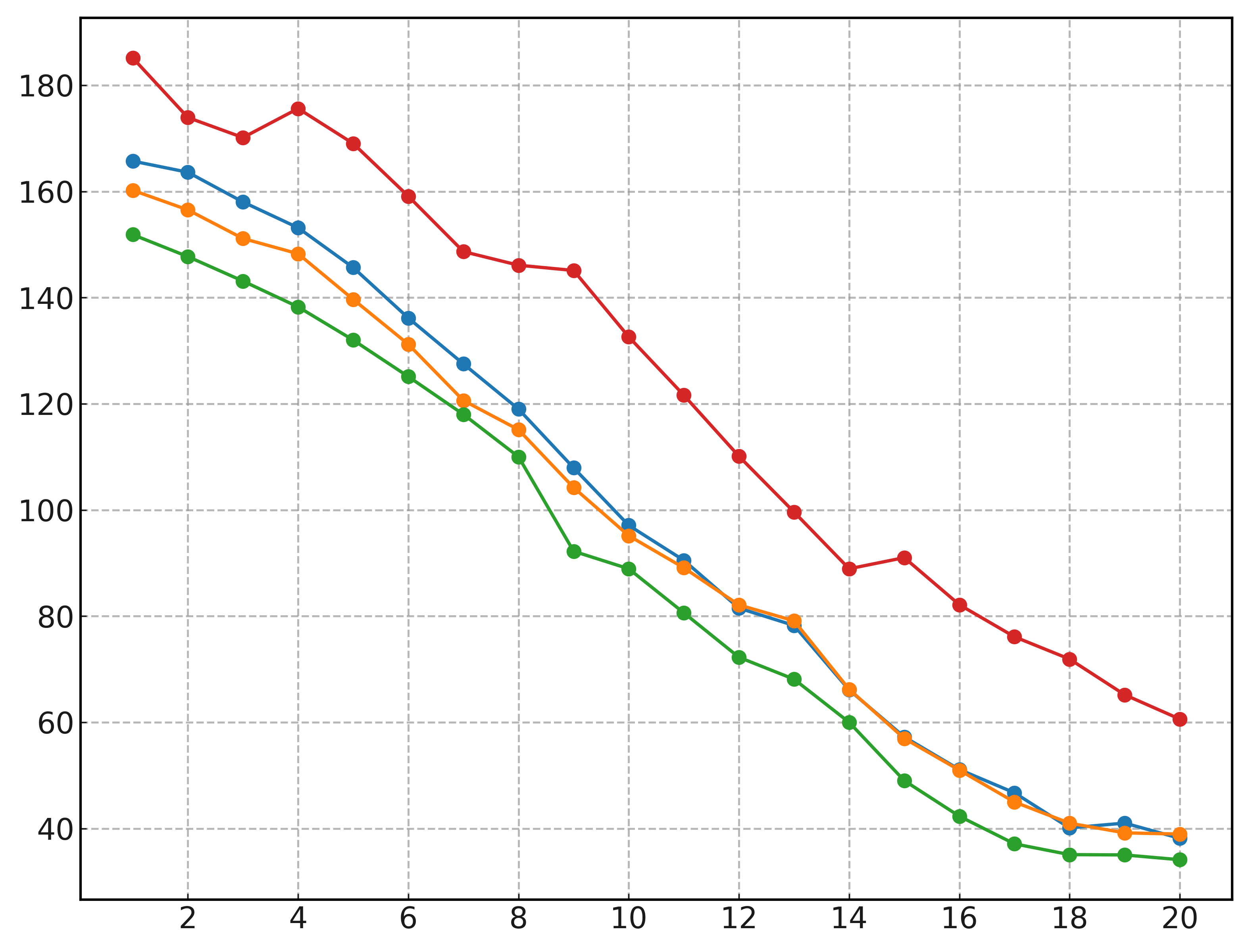}
    \end{subfigure}

    \begin{subfigure}[t]{0.5\textwidth}
        \centering
        \includegraphics[width=\textwidth]{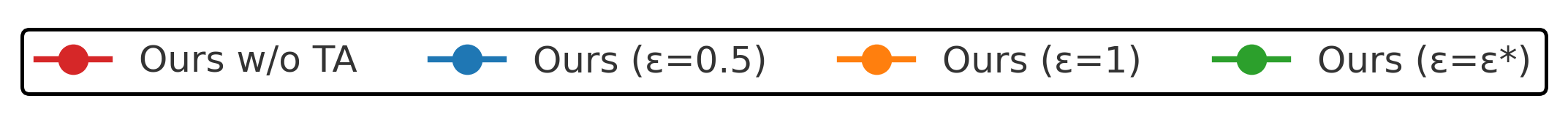}
    \end{subfigure}

    \caption{{\bf Visualization of the representation gap throughout training.} (a) FLD, (b) DLD, and (c) LLD plot the $L_2$ distance between teacher and student representations across epochs for different distillation settings. “Ours w/o TA” denotes multi-level distillation without a TA network. The X-axis represents epochs, and the Y-axis indicates $L_2$ distance.}
    \label{fig:ablationgraph}
\end{figure*}

\begin{table}[t]
    \begin{center}
    \setlength{\tabcolsep}{5pt}
    \renewcommand{\arraystretch}{0.9}
    \captionsetup{aboveskip=1pt, belowskip=1pt}
    \fontsize{9}{11}\selectfont   
    \begin{tabular}{cccccc}
    \toprule
    \multirow{2}{*}{Method} & \multirow{2}{*}{$\mathcal{L}_{FLD}$} & \multirow{2}{*}{$\mathcal{L}_{DLD}$} & \multicolumn{2}{c}{$\mathcal{L}_{LLD}$} & \multirow{2}{*}{\centering mIoU\(\uparrow\)(\%)} \\
    \cline{4-5}
    & & & $\mathcal{L}_{Base}^{L}$ & $\mathcal{L}_{Aux}^{L}$ \\
    \midrule 
    \multirow{2}{*}{Ours w/o TA} & \checkmark & \checkmark & \checkmark & - & 57.8 \\
    & \checkmark & \checkmark & \checkmark & \checkmark & 58.2 \\
    \midrule
    \multirow{4}{*}{Ours} & \checkmark & - & - & - & 58.5  \\
    & \checkmark & \checkmark & - & - &  59.4  \\
    & \checkmark & \checkmark & \checkmark & - &  60.5  \\
    & \cellcolor{gray!14}\checkmark & \cellcolor{gray!14}\checkmark & \cellcolor{gray!14}\checkmark & \cellcolor{gray!14}\checkmark & \cellcolor{gray!14}{\bf 60.8}  \\
    \bottomrule
    \end{tabular}
    \end{center}
    \caption{Ablation study of the TA network, multi-level distillation, and logit-level losses in BridgeTA.}
    \vspace{-15pt}
    \label{tab:KDablation}
\end{table}

\begin{figure*}
  \centering
    \includegraphics[width=0.99\linewidth]{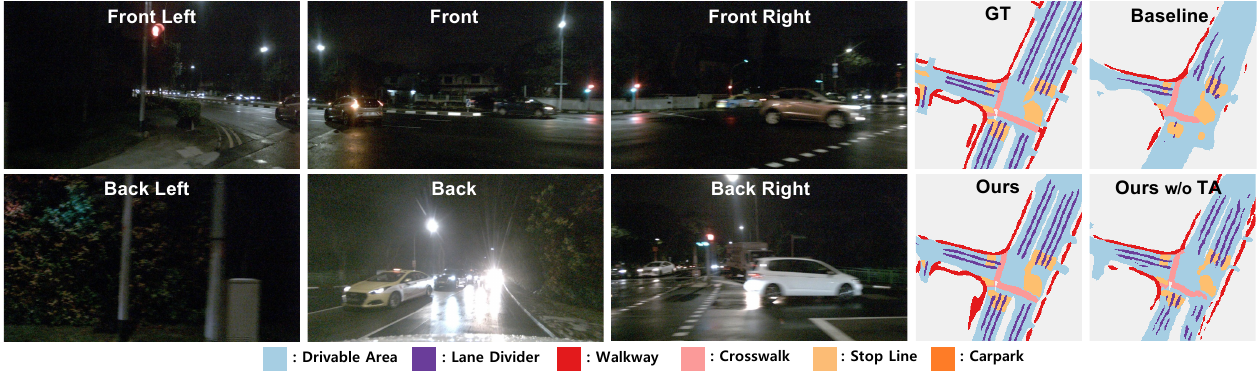} 
  \caption{ {\bf Qualitative results on nuScenes.} By effectively distilling geometric and structural information from the teacher, BridgeTA predicts essential road elements like Lane Dividers and Walkways more accurately than the Camera-only baseline even under challenging night-time conditions, demonstrating the effectiveness of the TA network.}
  \vspace{-5pt}
  \label{fig:qualitative}
\end{figure*}

\subsection{Ablation Studies}

\subsubsection{Effect of TA network and \textbf{$\varepsilon$} setting} After theoretically justifying the effectiveness of the TA network with Young’s inequality, we empirically validate this in ablation experiments. In Table~\ref{tab:KDablation}, Ours w/o TA, which excludes the TA network and applies only direct teacher-student distillation, achieves lower mIoU than our full approach. These results confirm that the TA network with dual-path distillation leads to more effective knowledge transfer.

We visualize the $L_2$ distance between teacher and student representations across all levels throughout training in Figure~\ref{fig:ablationgraph}. The results show that TA network enables faster, more stable convergence with consistently lower $L_2$ distances compared to Ours w/o TA. 

Furthermore, as shown in Figure~\ref{fig:ablationgraph}, applying the theoretically derived optimal weight $\varepsilon^*$ results in the smallest gap. This experimentally confirms that our mathematically determined value of $\varepsilon^*$ is indeed optimal for bridging the teacher-student representation gap.

\subsubsection{Multi-Level Distillation} Table~\ref{tab:KDablation} further shows that applying multi-level distillation terms, FLD, DLD, and LLD, to BridgeTA progressively boosts performance. Notably, incorporating all three levels simultaneously yields the best performance, which further highlights the complementary nature of each distillation term.

\subsubsection{Logit-Level Distillation (LLD)} We conducted ablations on the LLD loss components. As shown in the rightmost columns of Table~\ref{tab:KDablation}, combining both $\mathcal{L}_{Base}^L$ and $\mathcal{L}_{Aux}^L$ leads to the highest mIoU, indicating that both regression and distribution-based logit losses are essential.

\subsection{Qualitative Results}

We present the visualization of BEV map segmentation results in Figure~\ref{fig:qualitative} to highlight the effectiveness of BridgeTA and the role of the TA network in night-time scenarios. Since the baseline model relies solely on Camera inputs, it struggles to accurately predict essential road elements such as Lane Dividers and Walkways under low-light conditions. In contrast, BridgeTA leverages the fused LiDAR-Camera information from the teacher model to effectively distill both geometric and structural information. This enables the student to generate richer and more accurate feature representations even in night-time environments. This allows BridgeTA to deliver robust prediction results, significantly better than the baseline and Ours w/o TA. The results demonstrate the necessity and advantages of the TA network in handling challenging conditions, and BridgeTA consistently shows robust performance not only in night-time scenarios but also under other challenging driving scenarios.

%%%%%%%%%%%%%%%%%%%%%%%%%%%%%%%%%%%%%%%%%%%%%%%%%%%%%%%%%%%%%%%%%%%%%%%%%%%%%%%%
\section{Conclusions}
In this paper, we proposed BridgeTA, a distillation framework for BEV map segmentation that leverages a lightweight TA network to mitigate the representation gap between an LC fusion teacher and Camera-only student. Unlike prior work, BridgeTA preserves the student architecture with no additional inference cost. Our framework is theoretically grounded by Young’s inequality, which enables a principled decomposition of the distillation path into teacher-TA and TA-student with a tight alignment bound.

Extensive experiments demonstrate that BridgeTA effectively overcomes the representation gap and maintains robust performance even under adverse conditions. We believe our findings provide a promising direction for scalable and cost-effective BEV perception in autonomous driving.

\section*{Acknowledgements}
This work was supported by the KIST Institutional Program (No. 2E33801-25-015), by a grant from the Korea Evaluation Institute of Industrial Technology (KEIT), funded by the Korean government (MOTIE) (No. 20023455), and by the Autonomous Driving Center, Hyundai Motor Company.

%%%%%%%%%%%%%%%%%%%%%%%%%%%%%%%%%%%%%%%%%%%%%%%%%%%%%%%%%%%%%%%%%%%%%%%%%%%%%%%%

% \newpage
\bibliographystyle{IEEEtran}
\bibliography{ref}

@inproceedings{zhao2024simdistill,
  title={Simdistill: Simulated multi-modal distillation for bev 3d object detection},
  author={Zhao, Haimei and Zhang, Qiming and Zhao, Shanshan and Chen, Zhe and Zhang, Jing and Tao, Dacheng},
  booktitle={Proceedings of the AAAI Conference on Artificial Intelligence},
  volume={38},
  number={7},
  pages={7460--7468},
  year={2024}
}

@inproceedings{hao2024mapdistill,
  title={Mapdistill: Boosting efficient camera-based hd map construction via camera-lidar fusion model distillation},
  author={Hao, Xiaoshuai and Li, Ruikai and Zhang, Hui and Li, Dingzhe and Yin, Rong and Jung, Sangil and Park, Seung-In and Yoo, ByungIn and Zhao, Haimei and Zhang, Jing},
  booktitle={European Conference on Computer Vision},
  pages={166--183},
  year={2024},
  organization={Springer}
}

@inproceedings{zhou2023unidistill,
  title={UniDistill: A Universal Cross-Modality Knowledge Distillation Framework for 3D Object Detection in Bird's-Eye View},
  author={Zhou, Shengchao and Liu, Weizhou and Hu, Chen and Zhou, Shuchang and Ma, Chao},
  booktitle={Proceedings of the IEEE/CVF conference on computer vision and pattern recognition},
  pages={5116--5125},
  year={2023}
}

@article{ho2020ddpm,
  title={Denoising diffusion probabilistic models},
  author={Ho, Jonathan and Jain, Ajay and Abbeel, Pieter},
  journal={Advances in neural information processing systems},
  volume={33},
  pages={6840--6851},
  year={2020}
}

@inproceedings{caesar2020nuscenes,
  title={nuscenes: A multimodal dataset for autonomous driving},
  author={Caesar, Holger and Bankiti, Varun and Lang, Alex H and Vora, Sourabh and Liong, Venice Erin and Xu, Qiang and Krishnan, Anush and Pan, Yu and Baldan, Giancarlo and Beijbom, Oscar},
  booktitle={Proceedings of the IEEE/CVF conference on computer vision and pattern recognition},
  pages={11621--11631},
  year={2020}
}

@inproceedings{liu2023bevfusion,
  title={Bevfusion: Multi-task multi-sensor fusion with unified bird's-eye view representation},
  author={Liu, Zhijian and Tang, Haotian and Amini, Alexander and Yang, Xinyu and Mao, Huizi and Rus, Daniela L and Han, Song},
  booktitle={2023 IEEE international conference on robotics and automation (ICRA)},
  pages={2774--2781},
  year={2023},
  organization={IEEE}
}

@inproceedings{peng2023bevsegformer,
  title={Bevsegformer: Bird's eye view semantic segmentation from arbitrary camera rigs},
  author={Peng, Lang and Chen, Zhirong and Fu, Zhangjie and Liang, Pengpeng and Cheng, Erkang},
  booktitle={Proceedings of the IEEE/CVF Winter Conference on Applications of Computer Vision},
  pages={5935--5943},
  year={2023}
}

@inproceedings{pan2023baeformer,
  title={BAEFormer: Bi-Directional and Early Interaction Transformers for Bird's Eye View Semantic Segmentation},
  author={Pan, Cong and He, Yonghao and Peng, Junran and Zhang, Qian and Sui, Wei and Zhang, Zhaoxiang},
  booktitle={Proceedings of the IEEE/CVF Conference on Computer Vision and Pattern Recognition},
  pages={9590--9599},
  year={2023}
}

@article{li2024bevformer,
  title={Bevformer: learning bird's-eye-view representation from lidar-camera via spatiotemporal transformers},
  author={Li, Zhiqi and Wang, Wenhai and Li, Hongyang and Xie, Enze and Sima, Chonghao and Lu, Tong and Yu, Qiao and Dai, Jifeng},
  journal={IEEE Transactions on Pattern Analysis and Machine Intelligence},
  year={2024},
  publisher={IEEE}
}

@inproceedings{ren2022collaborative,
  title={Collaborative perception for autonomous driving: Current status and future trend},
  author={Ren, Shunli and Chen, Siheng and Zhang, Wenjun},
  booktitle={Proceedings of 2021 5th Chinese Conference on Swarm Intelligence and Cooperative Control},
  pages={682--692},
  year={2022},
  organization={Springer}
}

@article{zhao2024bev,
  title={BEV perception for autonomous driving: State of the art and future perspectives},
  author={Zhao, Junhui and Shi, Jingyue and Zhuo, Li},
  journal={Expert Systems with Applications},
  volume={258},
  pages={125103},
  year={2024},
  publisher={Elsevier}
}

@article{le2024diffuser,
  title={DifFUSER: Diffusion Model for Robust Multi-Sensor Fusion in 3D Object Detection and BEV Segmentation},
  author={Le, Duy-Tho and Shi, Hengcan and Cai, Jianfei and Rezatofighi, Hamid},
  journal={arXiv preprint arXiv:2404.04629},
  year={2024}
}

@inproceedings{kim2024broadbev,
  title={Broadbev: Collaborative lidar-camera fusion for broad-sighted bird’s eye view map construction},
  author={Kim, Minsu and Kim, Giseop and Jin, Kyong Hwan and Choi, Sunwook},
  booktitle={2024 IEEE International Conference on Robotics and Automation (ICRA)},
  pages={11125--11132},
  year={2024},
  organization={IEEE}
}

@inproceedings{borse2023x,
  title={X-Align: Cross-Modal Cross-View Alignment for Bird's-Eye-View Segmentation},
  author={Borse, Shubhankar and Klingner, Marvin and Kumar, Varun Ravi and Cai, Hong and Almuzairee, Abdulaziz and Yogamani, Senthil and Porikli, Fatih},
  booktitle={Proceedings of the IEEE/CVF Winter Conference on Applications of Computer Vision},
  pages={3287--3297},
  year={2023}
}

@article{chen2022gkt,
  title={Efficient and robust 2d-to-bev representation learning via geometry-guided kernel transformer},
  author={Chen, Shaoyu and Cheng, Tianheng and Wang, Xinggang and Meng, Wenming and Zhang, Qian and Liu, Wenyu},
  journal={arXiv preprint arXiv:2206.04584},
  year={2022}
}

@article{ge2023metabev,
  title={Metabev: Solving sensor failures for bev detection and map segmentation},
  author={Ge, Chongjian and Chen, Junsong and Xie, Enze and Wang, Zhongdao and Hong, Lanqing and Lu, Huchuan and Li, Zhenguo and Luo, Ping},
  journal={arXiv preprint arXiv:2304.09801},
  year={2023}
}

@article{xie2022m2bev,
  title={{M$^2$BEV}: Multi-Camera Joint 3D Detection and Segmentation with Unified Birds-Eye View Representation},
  author={Xie, Enze and Yu, Zhiding and Zhou, Daquan and Philion, Jonah and Anandkumar, Anima and Fidler, Sanja and Luo, Ping and Alvarez, Jose M},
  journal={arXiv preprint arXiv:2204.05088},
  year={2022}
}

@inproceedings{philion2020lift,
  title={Lift, splat, shoot: Encoding images from arbitrary camera rigs by implicitly unprojecting to 3d},
  author={Philion, Jonah and Fidler, Sanja},
  booktitle={Computer Vision--ECCV 2020: 16th European Conference, Glasgow, UK, August 23--28, 2020, Proceedings, Part XIV 16},
  pages={194--210},
  year={2020},
  organization={Springer}
}

@inproceedings{ji2023ddp,
  title={Ddp: Diffusion model for dense visual prediction},
  author={Ji, Yuanfeng and Chen, Zhe and Xie, Enze and Hong, Lanqing and Liu, Xihui and Liu, Zhaoqiang and Lu, Tong and Li, Zhenguo and Luo, Ping},
  booktitle={Proceedings of the IEEE/CVF International Conference on Computer Vision},
  pages={21741--21752},
  year={2023}
}

@article{woo2024location,
  title={Location-Aware Transformer Network for Bird's Eye View Semantic Segmentation},
  author={Woo, Suhan and Park, Minseong and Lee, Youngjo and Lee, Seoungwon and Kim, Euntai},
  journal={IEEE Transactions on Intelligent Vehicles},
  year={2025},
  publisher={IEEE}
}

@inproceedings{zhou2022cross,
  title={Cross-view transformers for real-time map-view semantic segmentation},
  author={Zhou, Brady and Kr{\"a}henb{\"u}hl, Philipp},
  booktitle={Proceedings of the IEEE/CVF conference on computer vision and pattern recognition},
  pages={13760--13769},
  year={2022}
}

@inproceedings{zhu2023mapprior,
  title={MapPrior: Bird's-Eye View Map Layout Estimation with Generative Models},
  author={Zhu, Xiyue and Zyrianov, Vlas and Liu, Zhijian and Wang, Shenlong},
  booktitle={Proceedings of the IEEE/CVF International Conference on Computer Vision},
  pages={8228--8239},
  year={2023}
}

@article{hinton2015distilling,
  title={Distilling the Knowledge in a Neural Network},
  author={Hinton, Geoffrey},
  journal={arXiv preprint arXiv:1503.02531},
  year={2015}
}

@inproceedings{chen2024residual,
  title={Residual Graph Convolutional Network for Bird's-Eye-View Semantic Segmentation},
  author={Chen, Qiuxiao and Qi, Xiaojun},
  booktitle={Proceedings of the IEEE/CVF Winter Conference on Applications of Computer Vision},
  pages={3324--3331},
  year={2024}
}

@misc{mmdet3d2020,
    title={{MMDetection3D: OpenMMLab} next-generation platform for general {3D} object detection},
    author={MMDetection3D Contributors},
    howpublished = {\url{https://github.com/open-mmlab/mmdetection3d}},
    year={2020}
}

@inproceedings{yang2022semantic,
  title={Cross-image relational knowledge distillation for semantic segmentation},
  author={Yang, Chuanguang and Zhou, Helong and An, Zhulin and Jiang, Xue and Xu, Yongjun and Zhang, Qian},
  booktitle={Proceedings of the IEEE/CVF Conference on Computer Vision and Pattern Recognition},
  pages={12319--12328},
  year={2022}
}

@inproceedings{liu2025monotakd,
  title={MonoTAKD: Teaching assistant knowledge distillation for monocular 3D object detection},
  author={Liu, Hou-I and Wu, Christine and Cheng, Jen-Hao and Chai, Wenhao and Wang, Shian-Yun and Liu, Gaowen and Latapie, Hugo and Wu, Jhih-Ciang and Hwang, Jenq-Neng and Shuai, Hong-Han and others},
  booktitle={Proceedings of the Computer Vision and Pattern Recognition Conference},
  pages={22266--22275},
  year={2025}
}

@article{li2022unifying,
  title={Unifying voxel-based representation with transformer for 3d object detection},
  author={Li, Yanwei and Chen, Yilun and Qi, Xiaojuan and Li, Zeming and Sun, Jian and Jia, Jiaya},
  journal={Advances in Neural Information Processing Systems},
  volume={35},
  pages={18442--18455},
  year={2022}
}

@article{chen2022bevdistill,
  title={Bevdistill: Cross-modal bev distillation for multi-view 3d object detection},
  author={Chen, Zehui and Li, Zhenyu and Zhang, Shiquan and Fang, Liangji and Jiang, Qinhong and Zhao, Feng},
  journal={arXiv preprint arXiv:2211.09386},
  year={2022}
}

@inproceedings{wang2024crosskd,
  title={CrossKD: Cross-head knowledge distillation for object detection},
  author={Wang, Jiabao and Chen, Yuming and Zheng, Zhaohui and Li, Xiang and Cheng, Ming-Ming and Hou, Qibin},
  booktitle={Proceedings of the IEEE/CVF conference on computer vision and pattern recognition},
  pages={16520--16530},
  year={2024}
}

@inproceedings{wang2023distillbev,
  title={Distillbev: Boosting multi-camera 3d object detection with cross-modal knowledge distillation},
  author={Wang, Zeyu and Li, Dingwen and Luo, Chenxu and Xie, Cihang and Yang, Xiaodong},
  booktitle={Proceedings of the IEEE/CVF International Conference on Computer Vision},
  pages={8637--8646},
  year={2023}
}

@inproceedings{bai2022transfusion,
  title={Transfusion: Robust lidar-camera fusion for 3d object detection with transformers},
  author={Bai, Xuyang and Hu, Zeyu and Zhu, Xinge and Huang, Qingqiu and Chen, Yilun and Fu, Hongbo and Tai, Chiew-Lan},
  booktitle={Proceedings of the IEEE/CVF conference on computer vision and pattern recognition},
  pages={1090--1099},
  year={2022}
}

@inproceedings{chen2023futr3d,
  title={Futr3d: A unified sensor fusion framework for 3d detection},
  author={Chen, Xuanyao and Zhang, Tianyuan and Wang, Yue and Wang, Yilun and Zhao, Hang},
  booktitle={proceedings of the IEEE/CVF conference on computer vision and pattern recognition},
  pages={172--181},
  year={2023}
}

@inproceedings{sindagi2019early1,
  title={Mvx-net: Multimodal voxelnet for 3d object detection},
  author={Sindagi, Vishwanath A and Zhou, Yin and Tuzel, Oncel},
  booktitle={2019 International Conference on Robotics and Automation (ICRA)},
  pages={7276--7282},
  year={2019},
  organization={IEEE}
}

@inproceedings{zhao2022logit2,
  title={Decoupled knowledge distillation},
  author={Zhao, Borui and Cui, Quan and Song, Renjie and Qiu, Yiyu and Liang, Jiajun},
  booktitle={Proceedings of the IEEE/CVF Conference on computer vision and pattern recognition},
  pages={11953--11962},
  year={2022}
}

@inproceedings{heo2019feature1,
  title={A comprehensive overhaul of feature distillation},
  author={Heo, Byeongho and Kim, Jeesoo and Yun, Sangdoo and Park, Hyojin and Kwak, Nojun and Choi, Jin Young},
  booktitle={Proceedings of the IEEE/CVF international conference on computer vision},
  pages={1921--1930},
  year={2019}
}

@inproceedings{liu2020residual,
  title={Residual feature distillation network for lightweight image super-resolution},
  author={Liu, Jie and Tang, Jie and Wu, Gangshan},
  booktitle={European conference on computer vision},
  pages={41--55},
  year={2020},
  organization={Springer}
}

@inproceedings{mirzadeh2020ta,
  title={Improved knowledge distillation via teacher assistant},
  author={Mirzadeh, Seyed Iman and Farajtabar, Mehrdad and Li, Ang and Levine, Nir and Matsukawa, Akihiro and Ghasemzadeh, Hassan},
  booktitle={Proceedings of the AAAI conference on artificial intelligence},
  volume={34},
  number={04},
  pages={5191--5198},
  year={2020}
}

@article{loshchilov2016sgdr,
  title={Sgdr: Stochastic gradient descent with warm restarts},
  author={Loshchilov, Ilya and Hutter, Frank},
  journal={arXiv preprint arXiv:1608.03983},
  year={2016}
}

@book{hardy1952inequalities,
  author    = {Hardy, G. H. and Littlewood, J. E. and {P{\'o}lya}, G.},
  title     = {Inequalities},
  publisher = {Cambridge University Press},
  year      = {1952},
}

\end{document}